\documentclass[10pt, conference, compsocconf]{IEEEtran}
\usepackage{cite}
\usepackage{gensymb}
\usepackage{graphicx,url,subfigure}

\usepackage[cmex10]{amsmath}
\begin{document}
%
\title{Character Keypoint-based Homography Estimation in Scanned Documents for Efficient Information Extraction}


\author{\IEEEauthorblockN{Kushagra Mahajan, Monika Sharma, Lovekesh Vig}
\IEEEauthorblockA{TCS Research, New Delhi, India\\
Email: \{kushagra.mahajan, monika.sharma1, lovekesh.vig\}@tcs.com}
}


%


\maketitle

\begin{abstract}
Precise homography estimation between multiple images is a pre-requisite for many computer vision applications. One application that is particularly relevant in today's digital era is the alignment of scanned or camera-captured document images such as insurance claim forms for information extraction. Traditional learning based approaches perform poorly due to the absence of an appropriate gradient. Feature based keypoint extraction techniques for homography estimation in real scene images either detect an extremely large number of inconsistent keypoints due to sharp textual edges, or produce inaccurate keypoint correspondences due to variations in illumination and viewpoint differences between document images. In this paper, we propose a novel algorithm for aligning scanned or camera-captured document images using character based keypoints and a reference template. The algorithm is both fast and accurate and utilizes a standard Optical character recognition (OCR) engine such as Tesseract to find character based unambiguous keypoints, which are utilized to identify precise keypoint correspondences between two images. Finally, the keypoints are used to compute the homography mapping between a test document and a template. We evaluated the proposed approach for information extraction on two real world anonymized datasets comprised of health insurance claim forms and the results support the viability of the proposed technique.



\end{abstract}
\vspace{-2mm}

%
\IEEEpeerreviewmaketitle

\section{Introduction}

Today's digital world calls for the digitization of every aspect of industry. One such aspect is the digitization of scanned or camera-captured document images such as bank receipts, insurance claim forms etc. for facilitating fast information retrieval from documents. Future references of `scanned' documents in the paper imply both scanned and camera-captured document images. Automating the task of information extraction from scanned documents suffers from difficulties arising due to variations in scanning of documents at different orientations and perspectives. This increases the likelihood of errors and additional human effort is required to extract relevant information from the documents. To circumvent this issue, we resort to image alignment techniques like homography estimation for aligning the given test document with a reference template document. Document alignment facilitates better performance by reducing the errors in field extraction and also reduces time and costs related to the digitization of scanned documents.

Homography estimation is essential for various tasks in computer vision like Simultaneous Localization and Mapping (SLAM), 3D reconstruction and panoramic image generation~\cite{mur2015orb, agarwal2009building, brown2007automatic}. A homography exists between projections of points on a 3D plane in two different views, i.e., a homography is said to be a transform / matrix which essentially converts points from one perspective to another perspective. To this end, we aim to find a transformation which allows matching / correspondence among pixels belonging to the test perturbed document and the template document image. This transformation can be represented as a matrix as shown in Equation~\ref{eqn:homography}.
\vspace{-2mm}
\begin{equation}
H = 
\begin{bmatrix}
    h_{11}       & h_{12} & h_{13} \\
    h_{21}       & h_{22} & h_{23} \\
    h_{31}       & h_{32} & h_{33} 
\label{eqn:homography}
\end{bmatrix}
\end{equation}
\vspace{-2mm}
\begin{equation}
    Y \sim HX \label{eqn:equation2}
\end{equation}

Here, the homography matrix $H$ has 8 free parameters as $h_{33} = 1$ or it imposes a unit vector constraint $(h_{11}^2 + h_{12}^2 + h_{13}^2 + h_{21}^2 + h_{22}^2 + h_{23}^2 + h_{31}^2 + h_{32}^2 + h_{33}^2 = 1)$. This means that we can compute a homography which describes how to transform the first set of points $X$ to the second set of points $Y$ using four pairs of matched points in our images.

\begin{figure}[h]
\begin{center}
\subfigure[]{\includegraphics[width=0.43\linewidth]{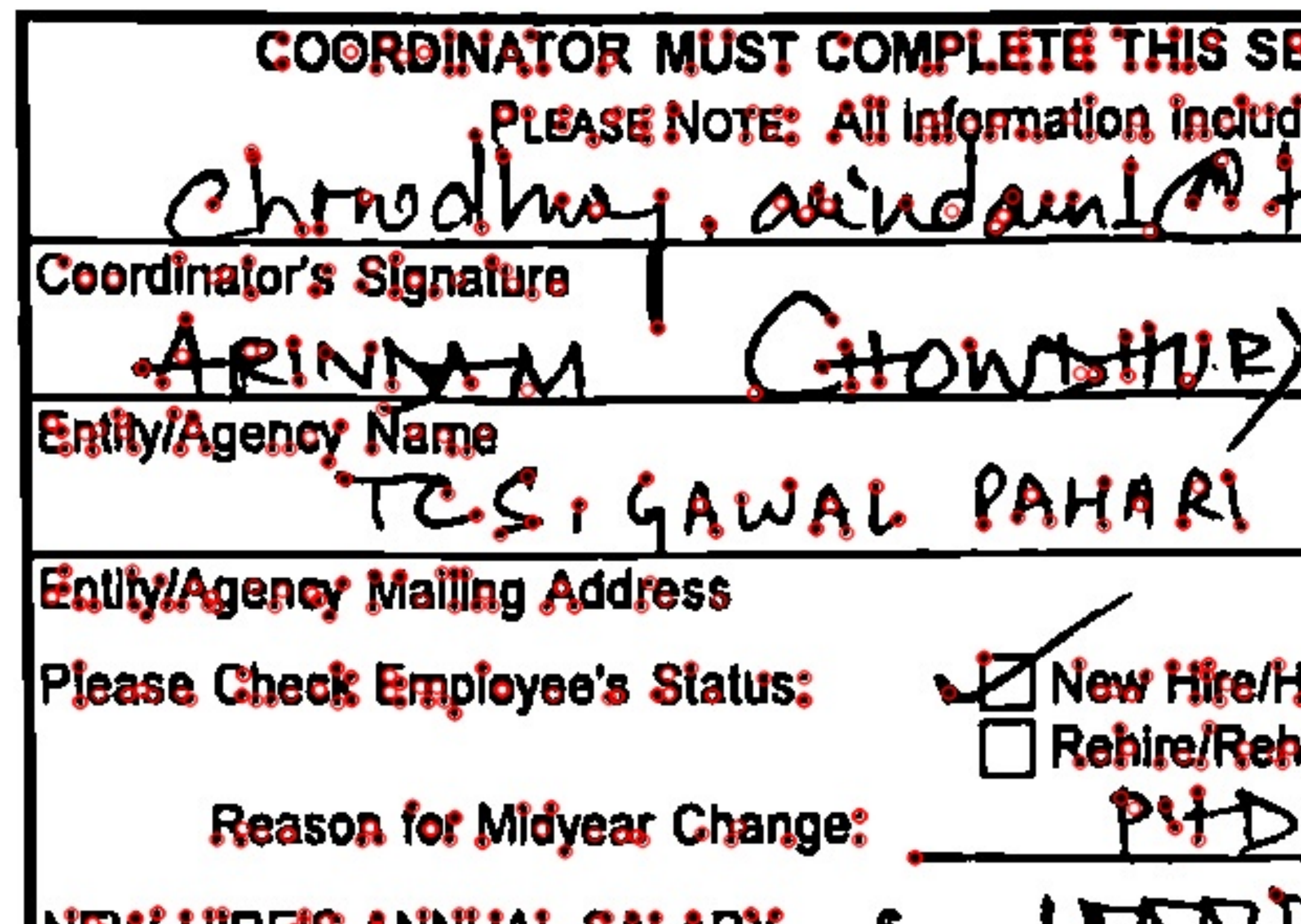}}
\subfigure[]{\includegraphics[width=0.43\linewidth, height = 26mm]{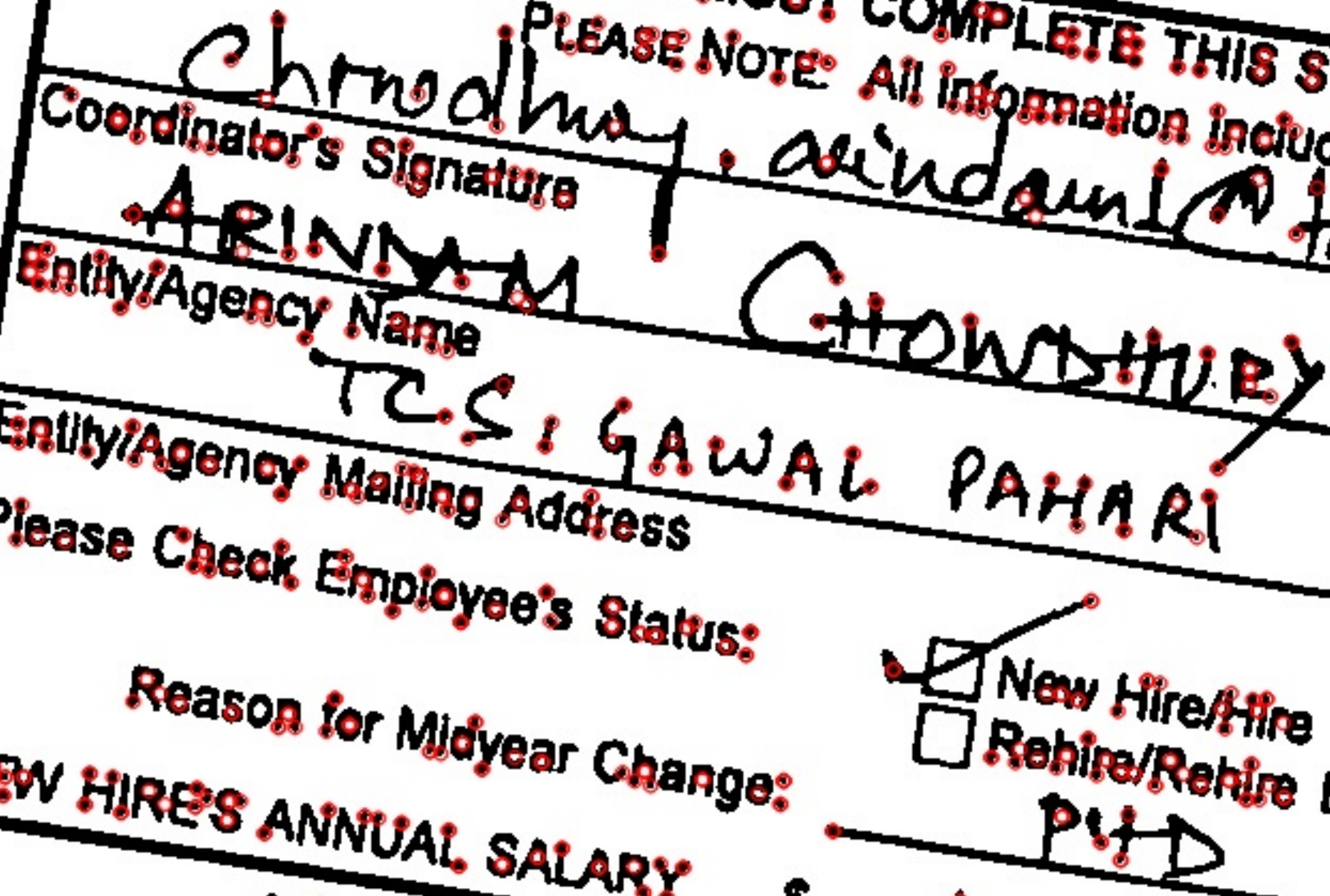}}\\
\subfigure[]{\includegraphics[width=0.43\linewidth]{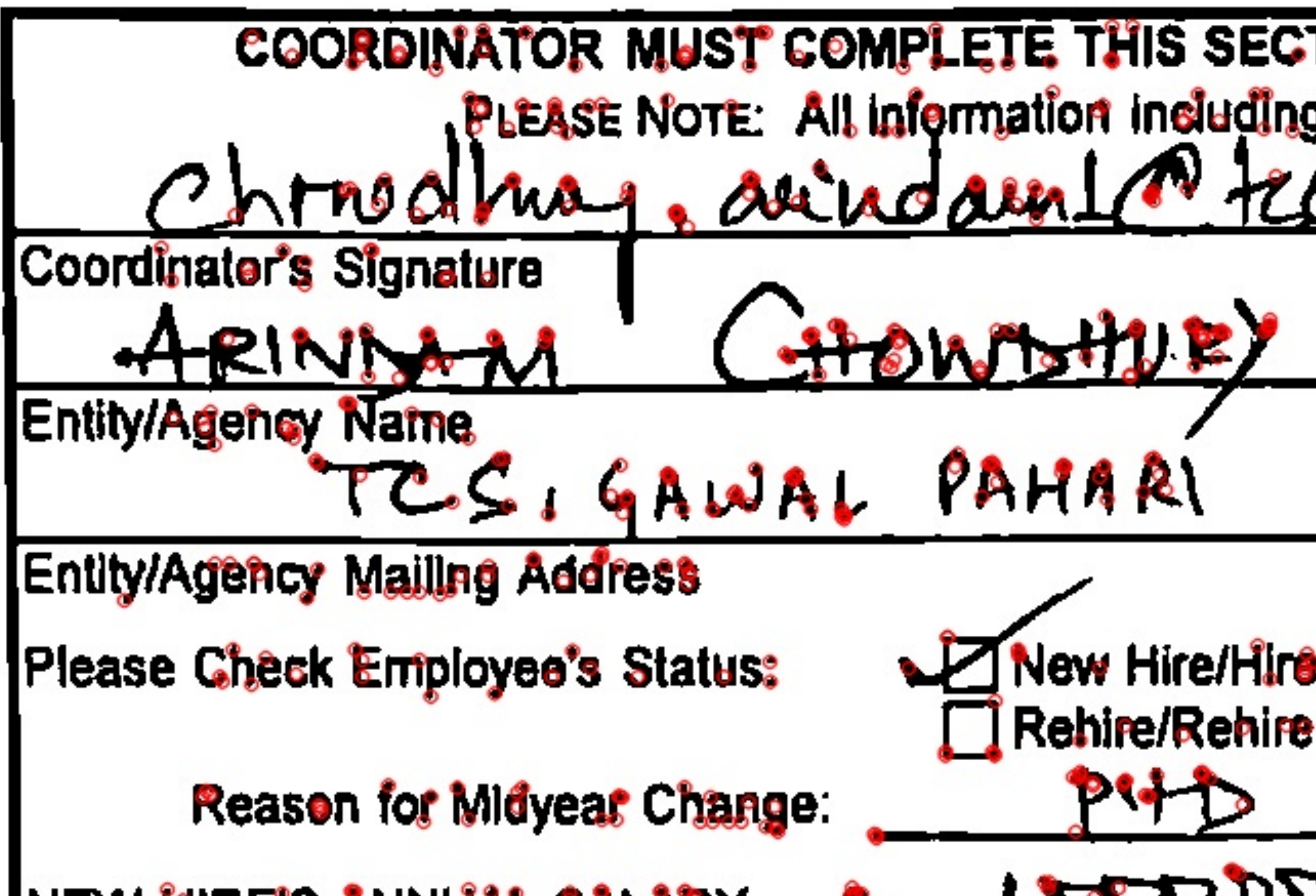}}
\subfigure[]{\includegraphics[width=0.43\linewidth, height = 25mm]{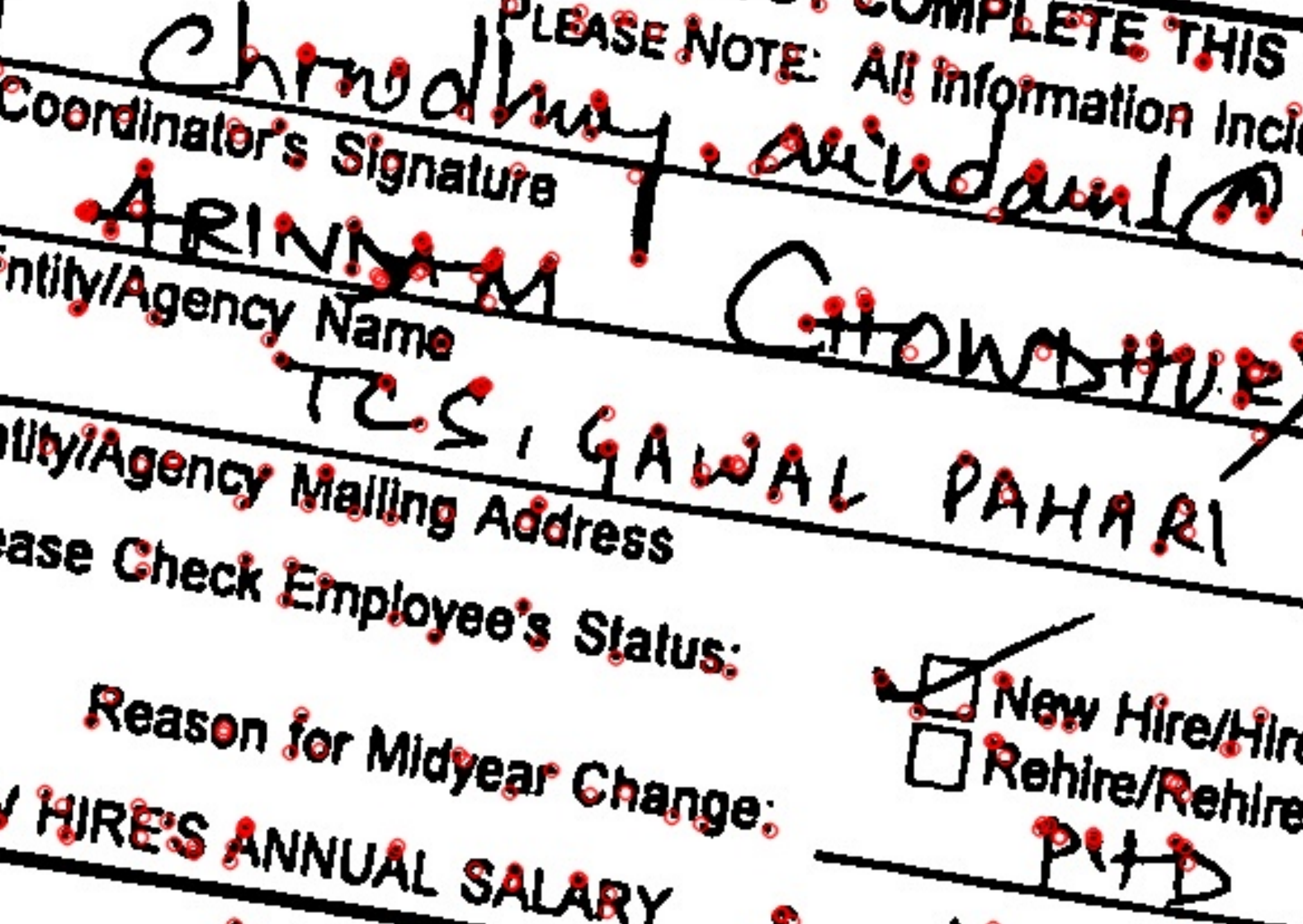}}
\end{center}
\vspace{-3mm}
\caption{(a) and (b) show SIFT keypoints extracted on patches of a document and a corresponding rotated document respectively. (c) and (d) show ORB~\cite{rublee2011orb} outputs on patches of the same image pair. Please note the difference in the keypoint detections between the original and rotated versions of the document. We observe that there is lack of consistency between the keypoint detections for both the feature descriptors i.e. very often, keypoints are not detected at corresponding locations in the two documents.}

\label{fig:motivation}
\end{figure}

Existing homography estimation techniques fall into two broad categories, namely \emph{direct pixel-based} and \emph{feature-based} methods. Among pixel based methods, Lucas Kanade's optical flow technique~\cite{lucas1981iterative} which utilizes the sum of squared differences between pixel intensity values as the error metric to estimate the motion of the pixels of the image contents is the most popular. An extension of the method was proposed by Lucey et al.~\cite{lucey2013fourier} which represents the images in complex 2D Fourier domain for improved performance. However, in the case of text document image alignment, these direct pixel based methods fail miserably to give the desired image alignment because sharp textual edges do not provide a smooth gradient which can be used for learning the homography. Feature-based methods first extract the keypoints, and then match the corresponding keypoints between the original and transformed images using their respective keypoint descriptors. This keypoint correspondence is used to estimate the homography between the two images. The most fundamental feature-descriptors used for keypoint extraction and matching tasks are Scale-Invariant Feature Transform (SIFT)~\cite{lowe2004distinctive} and Oriented FAST and Rotated BRIEF (ORB)~\cite{rublee2011orb}. When we use these feature-descriptors for detecting keypoints in the text document images, a large number of inconsistent keypoints are detected due to sharp textual edges producing inaccurate keypoint correspondences
, as illustrated in Figure~\ref{fig:motivation}.


To overcome these challenges, in this paper, we propose a novel and robust algorithm for aligning scanned document images using character based keypoints and a reference empty template document. The proposed method utilizes a standard OCR such as Tesseract~\cite{tesseract} to find unambiguous character based keypoints, which are subsequently utilized to identify precise keypoints correspondences between the two document images. Finally, the keypoint correspondences are used to compute the homography mapping between the two document images. The proposed approach is fast, robust and involves minimal memory requirements in contrast to complex deep learning based approaches which require huge memory to store the model and large compute power to produce results even during test time. Hence, this method is ideal for real-time computation on mobile devices and other electronic gadgets with resource constraints.

\begin{figure*}[h]
  \centering
  \includegraphics[width=0.7\linewidth]{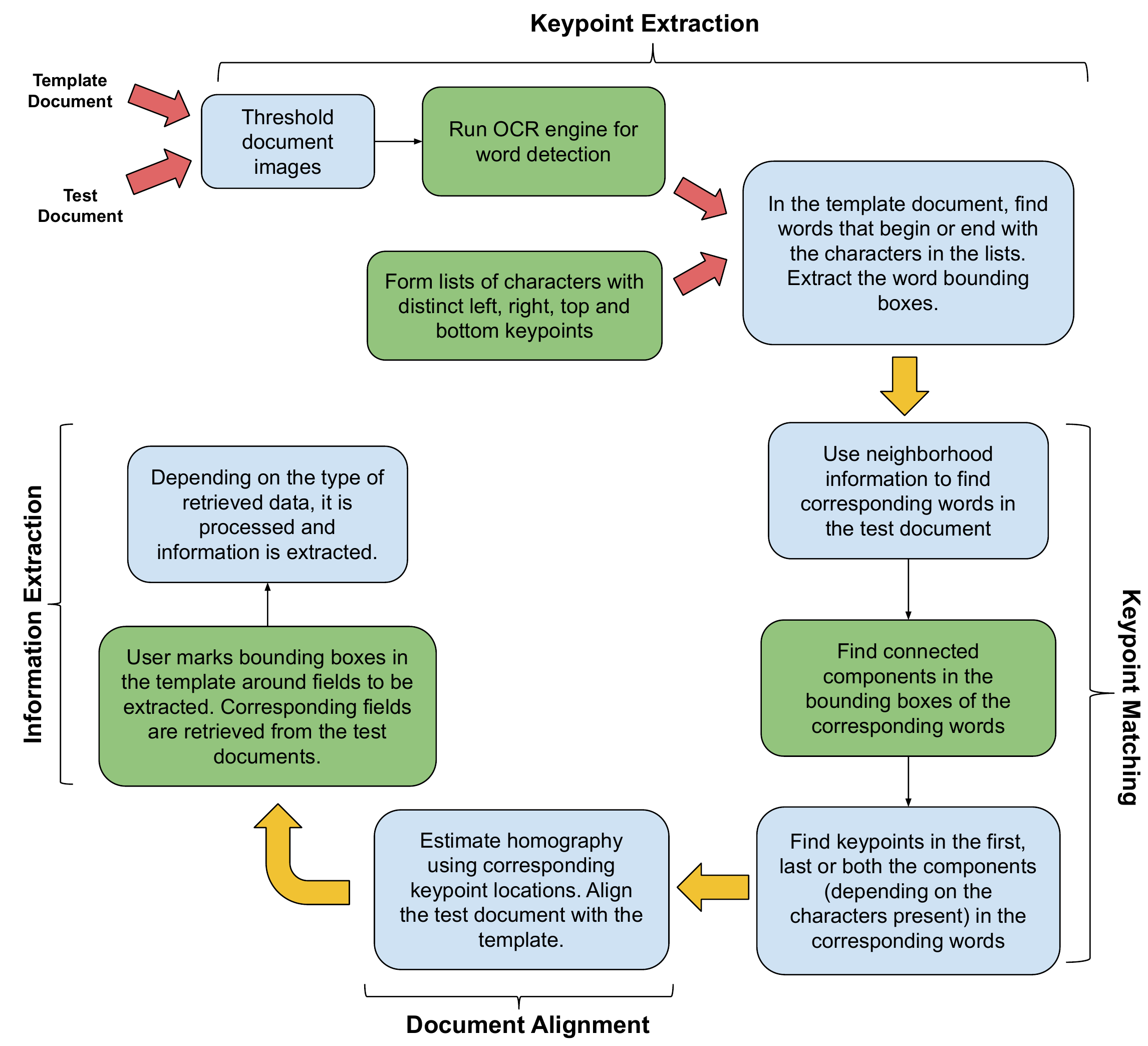}
  \vspace{-3mm}
  \caption{Flowchart showing the entire pipeline for information extraction from scanned document images after aligning with the template document using character keypoint-based homography estimation.}
  \label{fig:flowchart}
\end{figure*}

To summarize, our contributions in the paper are as follows :
\begin{itemize}
    \item We propose a novel, fast and memory efficient algorithm for robust character based unambiguous keypoint detection, extracted using a standard OCR like Tesseract, from scanned textual documents.
    \item We demonstrate how existing homography estimation approaches perform poorly when the problem space is extended to scanned document images. The limitations of these approaches are analyzed to come up with our methodology. 
    \item We show the effectiveness of our proposed approach using information extraction from two real world anonymized datasets comprised of health insurance claim forms, and present the qualitative and quantitative results in Section~\ref{sec:experimental-results}.
\end{itemize}

Remainder of the paper is organized as follows : Section~\ref{sec:related-work} discusses some of the prior work done in the field of image alignment, keypoint extraction and homography estimation. A detailed step-by-step explanation of the proposed approach is presented in Section~\ref{sec:proposed-approach}. Section~\ref{sec:dataset} gives details of the two real world anonymized health insurance claim form datasets used for document alignment. Subsequently, the experimental results and discussions on the same are given in Section~\ref{sec:experimental-results}. Finally, we conclude the paper in Section~\ref{sec:conclusion}.

\vspace{0mm}
\subsection{Related Work}
\label{sec:related-work}
By far, feature based methods relying on detection and matching of local image features are the most widely used techniques for homography estimation and subsequent image alignment. ~\cite{takeda2011real} uses the centroids of words in the document to compute the features. Since centroid computation at different orientations suffers from lack of precision, these features cannot be used for our task which requires exactness. ~\cite{block2007sitt} used structures in the text document like punctuation characters as keypoints for document mosaicing, while Royer et al.~\cite{royer2017benchmarking} explored keypoint selection methods which reduce the number of extracted keypoints for improved document image matching. Recently, deep neural networks have become popular to obtain powerful feature descriptors~\cite{zagoruyko2015learning, simo2015discriminative, han2015matchnet, balntas2016pn} compared with the traditional descriptors. These approaches create patches with descriptors computed for each patch. Similarity scores and distance measures between the descriptors are then used for obtaining the matches. Similarly, ~\cite{rocco2017convolutional} proposed an end-to-end architecture for learning affine transformations without manual annotation where features for each of the two images are extracted through a siamese architecture, followed by trainable matching and geometric parameter estimation producing state-of-the-art results on the Proposal Flow dataset. DeTone et al.~\cite{detone2016deep} devised a deep neural network to address the problem of homography estimation. They estimate the displacements between the four corners of the original and perturbed images in a supervised manner, and map it to the corresponding homography matrix. Another work of particular interest to us is done by Nguyen et al.~\cite{nguyen2018unsupervised} which trains a Convolutional Neural Network (CNN) for unsupervised learning of planar homographies, achieving faster inference and superior performance compared to the supervised counterparts. 


\section{Proposed Approach}
\label{sec:proposed-approach}
In this section, we discuss in detail the proposed method for information extraction from documents aligned using character keypoint-based homography estimation. The task of information extraction in a scanned document image involves finding values of fields of interest marked by a user. To accomplish this, the proposed method requires a template document image with empty fields for each document dataset. We attempt to align the test document images which have filled fields of interest with the empty template document. After the documents are aligned, the desired text fields are retrieved from the filled test documents, and information is read using an Optical Character Recognition (OCR) engine~\cite{tesseract} and Handwritten Text Recognition (HTR) deep network~\cite{chowdhury2018efficient}. The entire pipeline for the algorithm is shown in Figure~\ref{fig:flowchart}.

\vspace{0mm}
\subsection{Character based Keypoint Extraction}
\label{subsec:keypoint-extraction}
We begin by thresholding the empty template document as well as the filled test document from which information is to be extracted. Thresholding allows us to mitigate the impact of illumination variations present in scanned document images. Next, we use Tesseract as the OCR on both the documents, read all the words present in them and return the coordinates of the bounding boxes for each word. We observe an inherent trait present in certain characters like 'A', 'T', 'r' etc. that they have distinct tips. This attribute of characters is used to extract precise and unambiguous character keypoints.

\begin{figure}
\centering
  \includegraphics[width=0.8\linewidth]{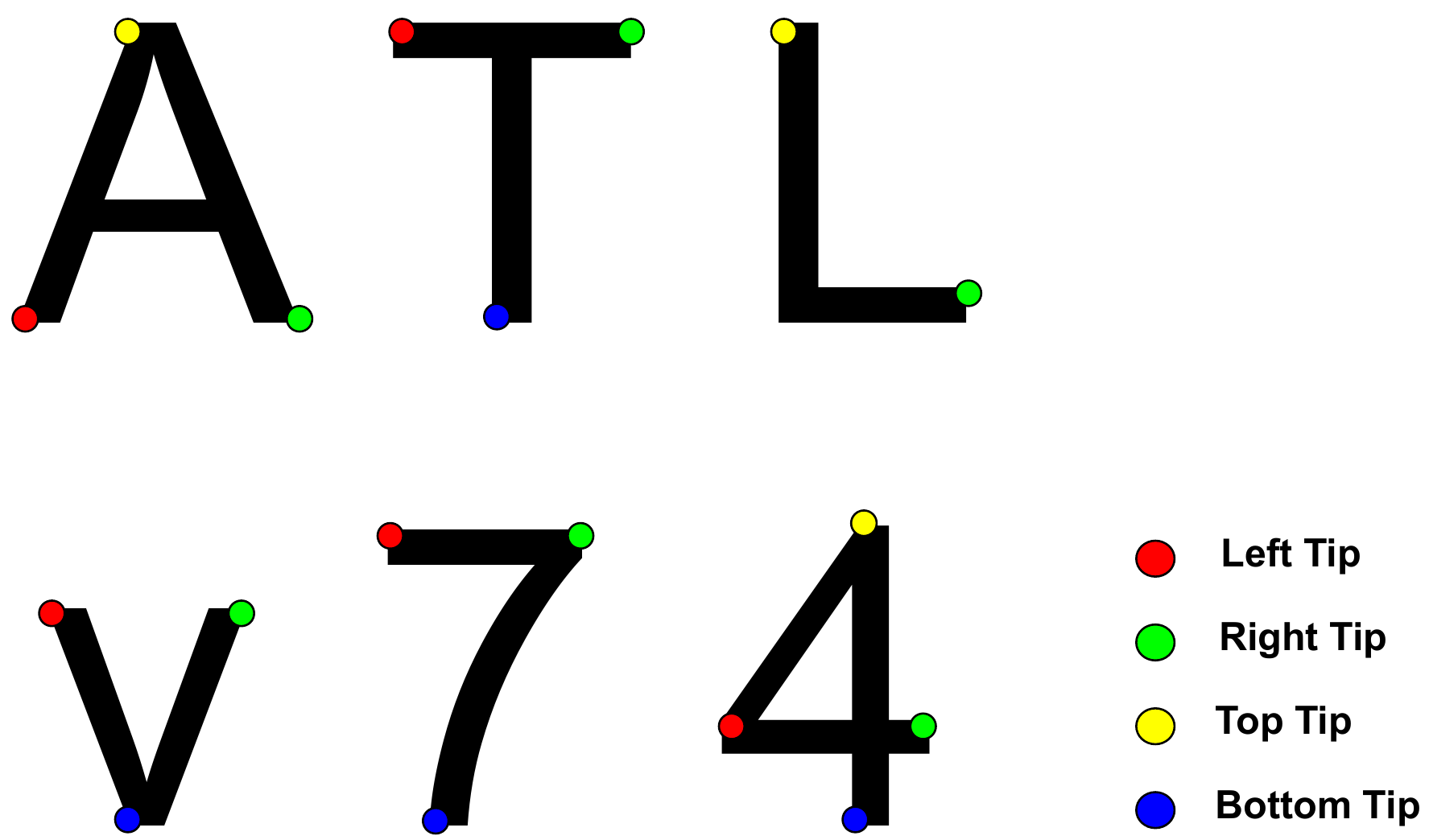}
  \vspace{-2mm}
  \caption{Left, right, top and bottom tips are shown for some of the characters included in the ${begCharList}$, ${endCharList}$, ${topCharList}$ and ${bottomCharList}$ respectively.}
  \label{fig:character_keypoints}
\end{figure}

We create four separate lists, namely ${begCharList}$, ${endCharList}$, ${topCharList}$, ${bottomCharList}$ for characters that have distinct tips on the left, right, top or bottom respectively, as shown in Figure~\ref{fig:character_keypoints}. For example, ${begCharList}$ includes characters such as 'A', 'V', 'T', 'Y', '4', 'v', 'w' etc. with a distinct left tip, and ${endCharList}$ consists of characters like 'V', 'T', 'L', 'Y', '7', 'r' etc. with a distinct right tip. We refrain from selecting characters like 'O' or 'D' since there is impreciseness in the keypoint detection in the curved portions which can ultimately impact the overall image alignment. We ensure that the accuracy of our system is not compromised, therefore, only unambiguous characters are considered for keypoint detection.

In the next step, we extract the word patch from the document that either begins with one of the characters present in ${begCharList}$, ${topCharList}$ or ${bottomCharList}$, or ends with one of the characters in ${endCharList}$, ${topCharList}$ or ${bottomCharList}$. After that, we run the connected components algorithm to find the components for that word. Then, we look at the leftmost and the rightmost component in the word. We search for the distinct tips in the first component if the component character is in ${begCharList}$, ${topCharList}$ or ${bottomCharList}$, and the last component if the component character is in ${endCharList}$, ${topCharList}$ or ${bottomCharList}$. As a result, we get a set of keypoints in the template document and the corresponding keypoints in the test document. We only use the first and last components of the word since these are guaranteed to include the first and last characters. Ideally each character should be detected as a separate component. However, in reality, this may not be the case because characters within the word may touch each other as a result of thresholding. 

To improve the performance of our proposed method, certain heuristic checks are also imposed. Words with two or lesser characters are ignored since they are more likely to be false positive detections by Tesseract. A constraint is put on the font size. This is done because it was found empirically that very small font sizes tend to get broken during thresholding and are likely to be incorrectly detected by Tesseract. We use the \emph{Enchant} spell checking library in Python to make sure that the words used for detecting the keypoints are valid words of the English language. This prevents any junk words from being used for keypoint detection since these words might be detected differently across the template and test documents. \\


\vspace{-3mm}
\subsection{Keypoint Matching}
\label{subsec:keypoint-matching}
The next step is to obtain correspondences between keypoints of the template and test documents. Since a word in a template can appear multiple times, we need to be sure that keypoints of the corresponding words are being matched. For this, we take a neighbourhood region centred at the word under consideration in the template document. A similar region is taken around the matched candidate word in the test document. In an ideal scenario, all the words in the template word neighbourhood region should also occur in the test candidate neighbourhood region. However, the test candidate neighbourhood region can have some additional words in the form of handwritten or printed text from the filled fields. So, we keep a threshold of 90\%, which means that if the test candidate neighbourhood has at least 90\% of the words present in the template word neighbourhood, then the test candidate is the corresponding matching word in the test document. An analogy can be drawn with feature matching involving commonly used local descriptors like SIFT and ORB which compute keypoint descriptors using a neighbourhood region around the keypoint, and use the similarity between descriptors for the matching task.

\renewcommand{\tabcolsep}{0.5mm}
\begin{table}[h]
\begin{center}
\vspace{-2mm}
\caption{Character recognition accuracy for fields in the first insurance dataset. Column (a) gives the accuracy of the printed text, column (b) shows the accuracy for handwritten text tested on the HTR~\cite{chowdhury2018efficient}, while column (c) mentions the accuracy of handwritten text using the Google Vision API.}
\vspace{1mm}
\begin{tabular}{|c|c|c|c|}
\hline
\textbf{Field} & {\textbf{Tesseract (A)}} & {\textbf{HTR~\cite{chowdhury2018efficient}(B)}} & \textbf{Vision API (C)} \\
\hline
\textbf{Name} & 98.6\% & 88\% & 92.2\% \\
\textbf{Pet Name} & 99.2\% & 89.5\% & 92.9\%\\
\textbf{Address} & 98.3\% & 80.4\% & 85.8\%\\
\textbf{Hospital} & 98.7\% & 77.3\% & 82.5\%\\
\textbf{Injury} & 97.1\% & 78\% & 82.6\%\\
\hline
\end{tabular}
\end{center}
\label{tb:nationwide_character}
\end{table}

\vspace{-3mm}
\subsection{Document Alignment}
The next step in the pipeline is to find the homography mapping between the template and test documents from the keypoint correspondences obtained in the previous step. OpenCV's \emph{findHomography} method finds this transformation matrix between the documents. The method makes use of Equations~\ref{eqn:homography} and \ref{eqn:equation2} to find the transformation matrix $H$. 
The noise in keypoint detection might hamper system performance. To make the method more robust, we supply a much larger set of keypoint pairs than the minimum four required for homography estimation. RANSAC~\cite{ransac} is used to get rid of any noise in the system which appears in the form of outliers. The transformation obtained is then applied to the test document using \emph{warpPerspective} function in OpenCV which takes the transformation matrix and the image on which the transformation matrix is to be applied as input. This operation is equivalent to Equation~\ref{eqn:equation2} being applied to every pixel in the test document. It gives us the test document aligned with the template document. \\


\vspace{-4mm}
\subsection{Information Extraction}
Having aligned the test document with the template, the user now simply marks the text field regions in the template document that need to be extracted from each of the test documents. The corresponding patches in the test documents are retrieved. Textual information is best read if nature of the text is known. Hence, we train a convolutional neural network based classifier to identify whether a textual field is handwritten or printed. The classifier gives near-perfect performance, with the accuracy being 98.5\%. Now, if the text is recognized as printed, the retrieved field patch is sent to Tesseract for recognition. For handwritten text, we use the work of Chowdhury et al.~\cite{chowdhury2018efficient} and the Google Vision API~\footnote{Google Cloud Vision Api : https://cloud.google.com/vision/} for recognition.


\section{Dataset}
\label{sec:dataset}
We evaluated our proposed approach on two real world anonymized document datasets. The first dataset consists of 15 insurance claim forms and one corresponding empty template form. The second dataset contains 15 life insurance application forms along with one corresponding empty template form. This dataset does not have filled text in printed form. The filled data is only in the form of handwritten text. These datasets contain documents with variations in illumination, different backgrounds like wooden table and also, the documents are affine transformed relative to the template document. All the documents are resized to $1600 \times 2400$, and converted to grayscale for further experiments. 


\begin{figure}[h]
\begin{center}
\subfigure[]{\includegraphics[width=0.23\linewidth]{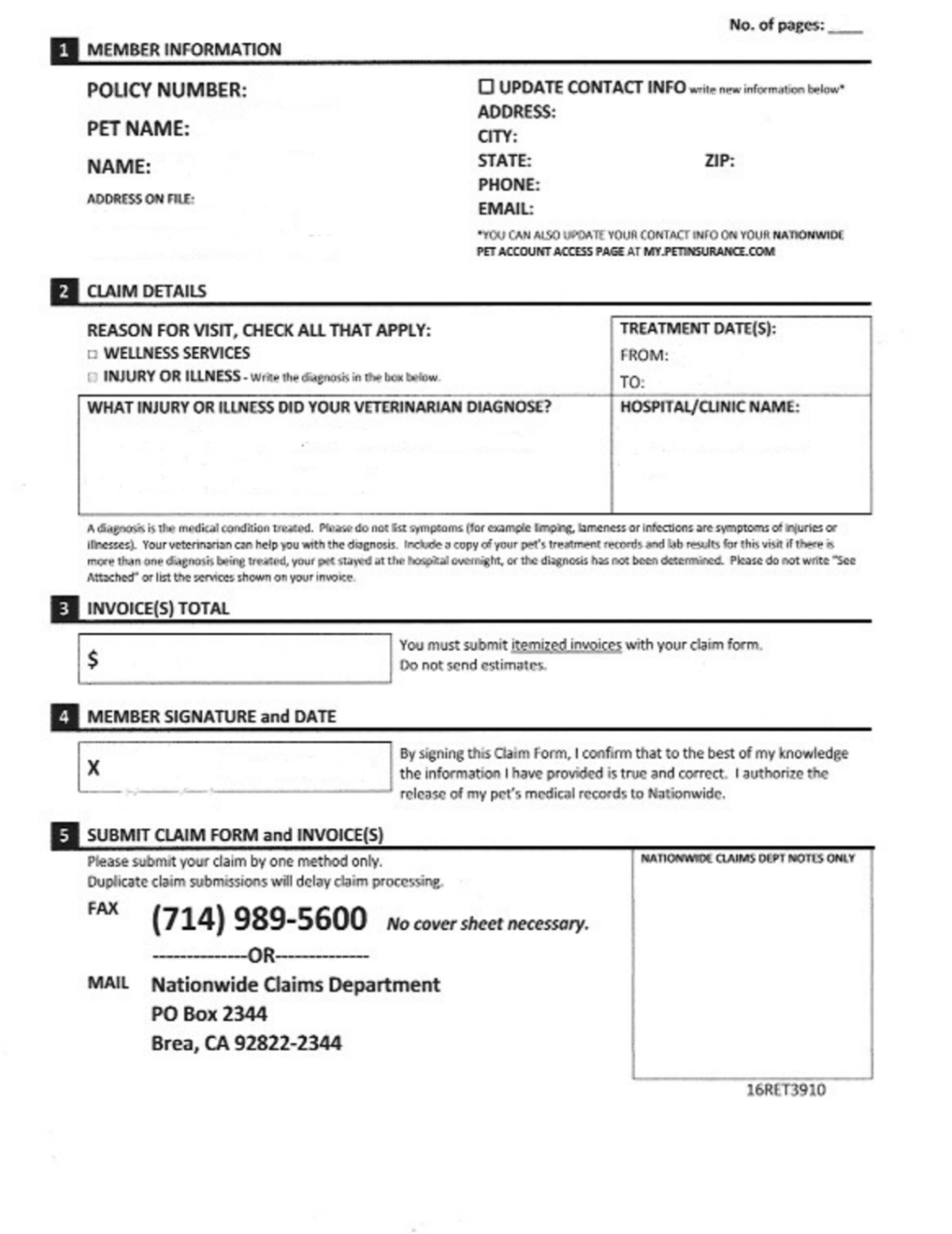}}
\subfigure[]{\includegraphics[width=0.23\linewidth]{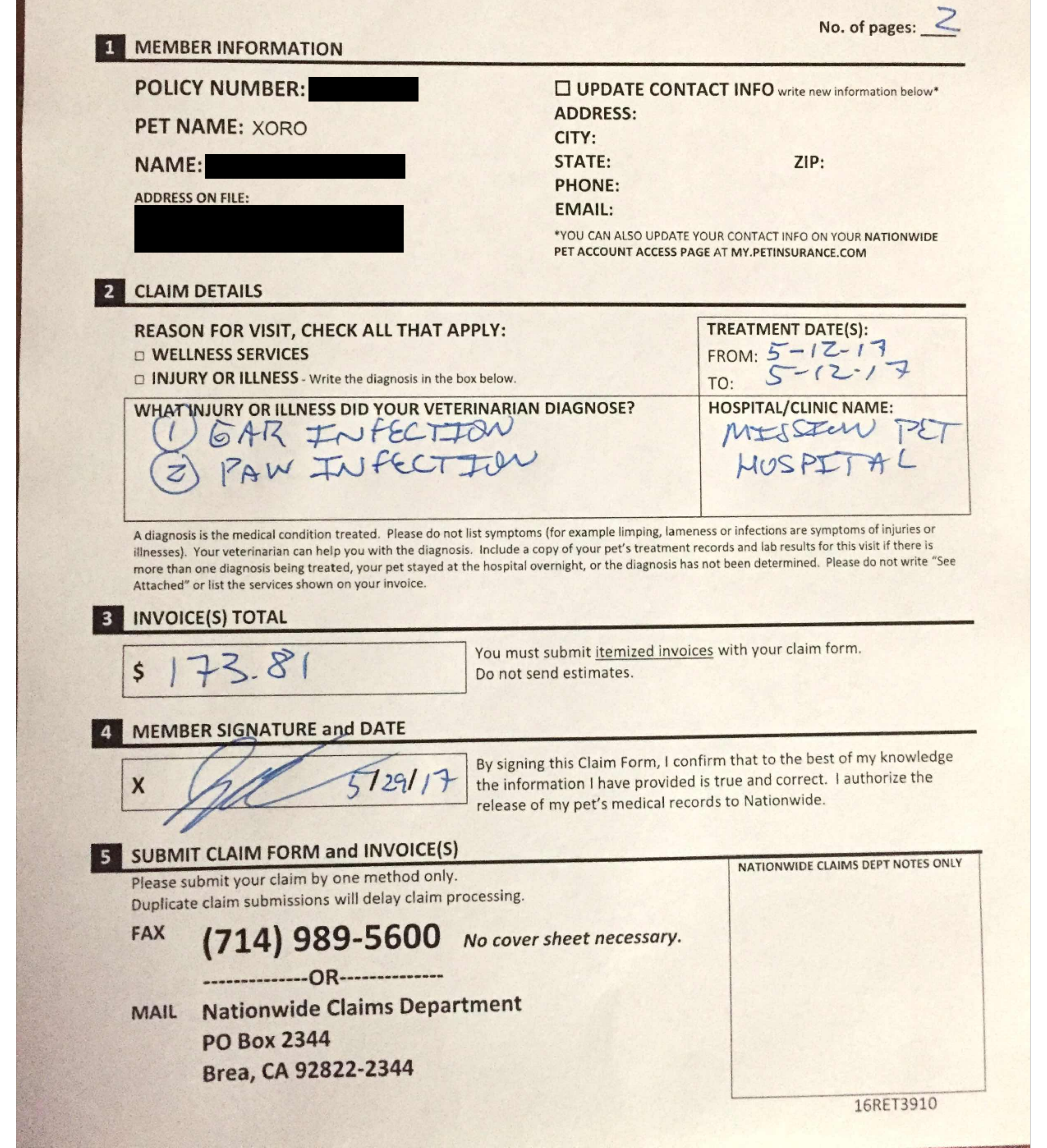}}
\subfigure[]{\includegraphics[width=0.23\linewidth]{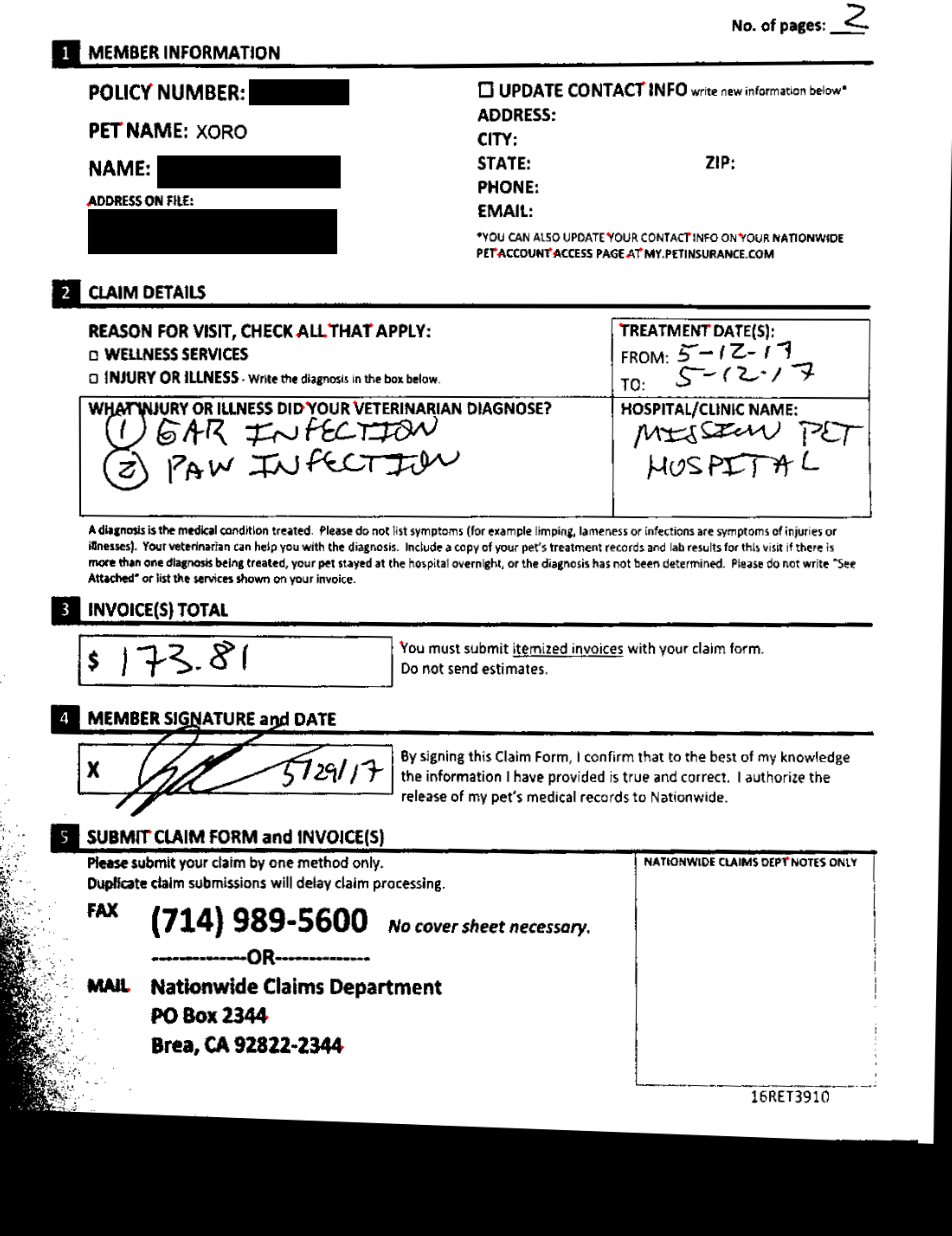}}
\subfigure[]{\includegraphics[width=0.23\linewidth]{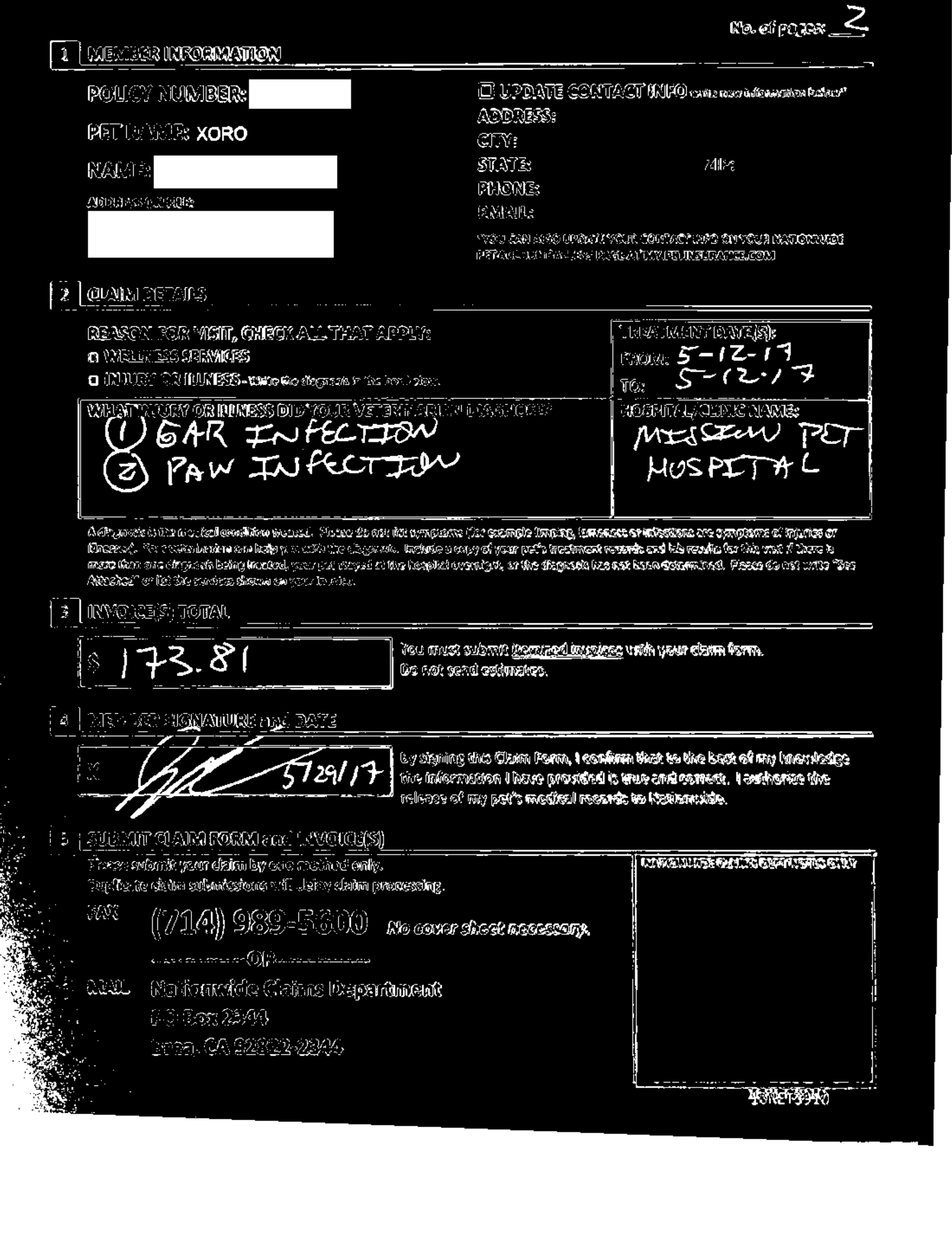}}\\
\subfigure[]{\includegraphics[width=0.23\linewidth]{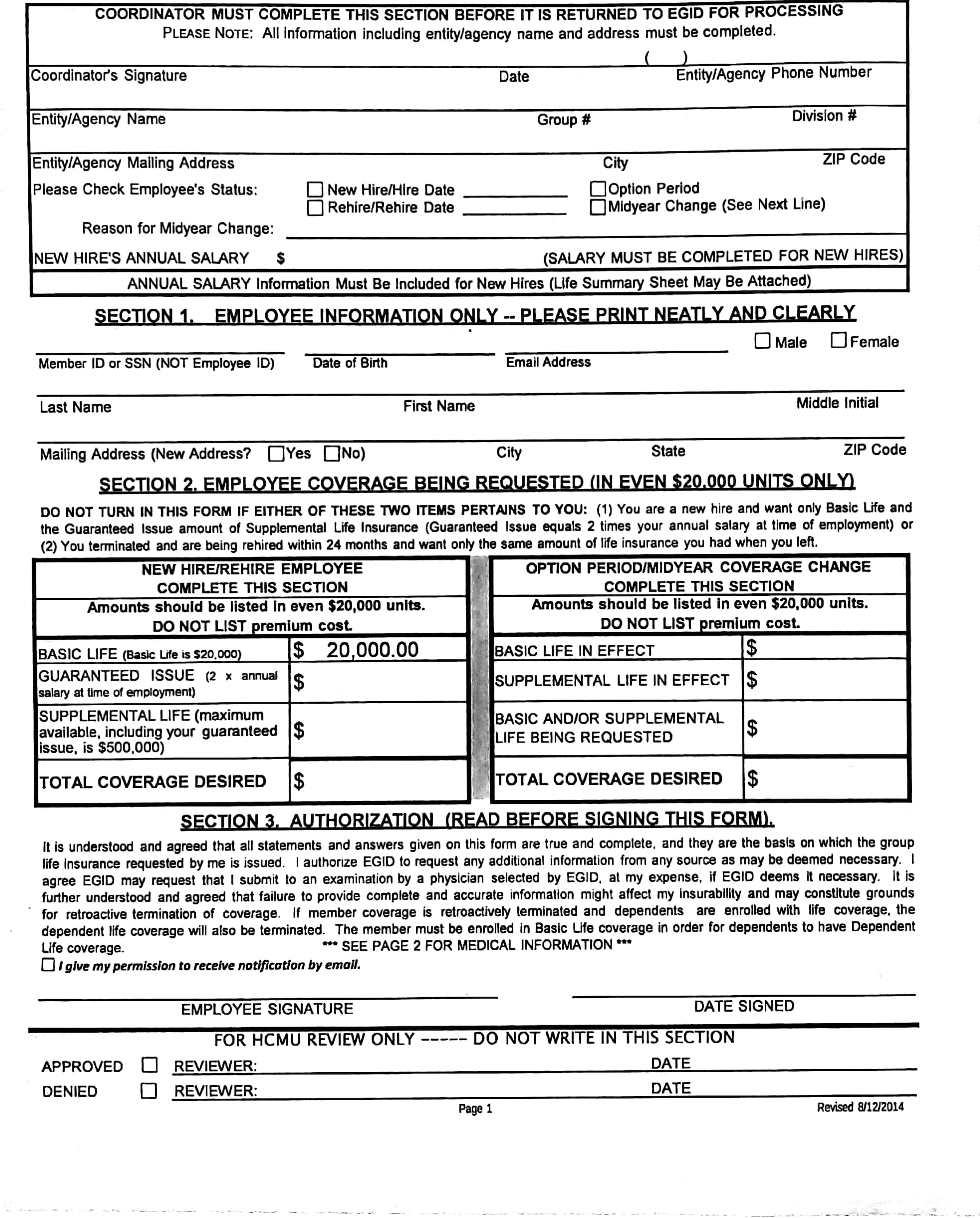}}
\subfigure[]{\includegraphics[width=0.23\linewidth]{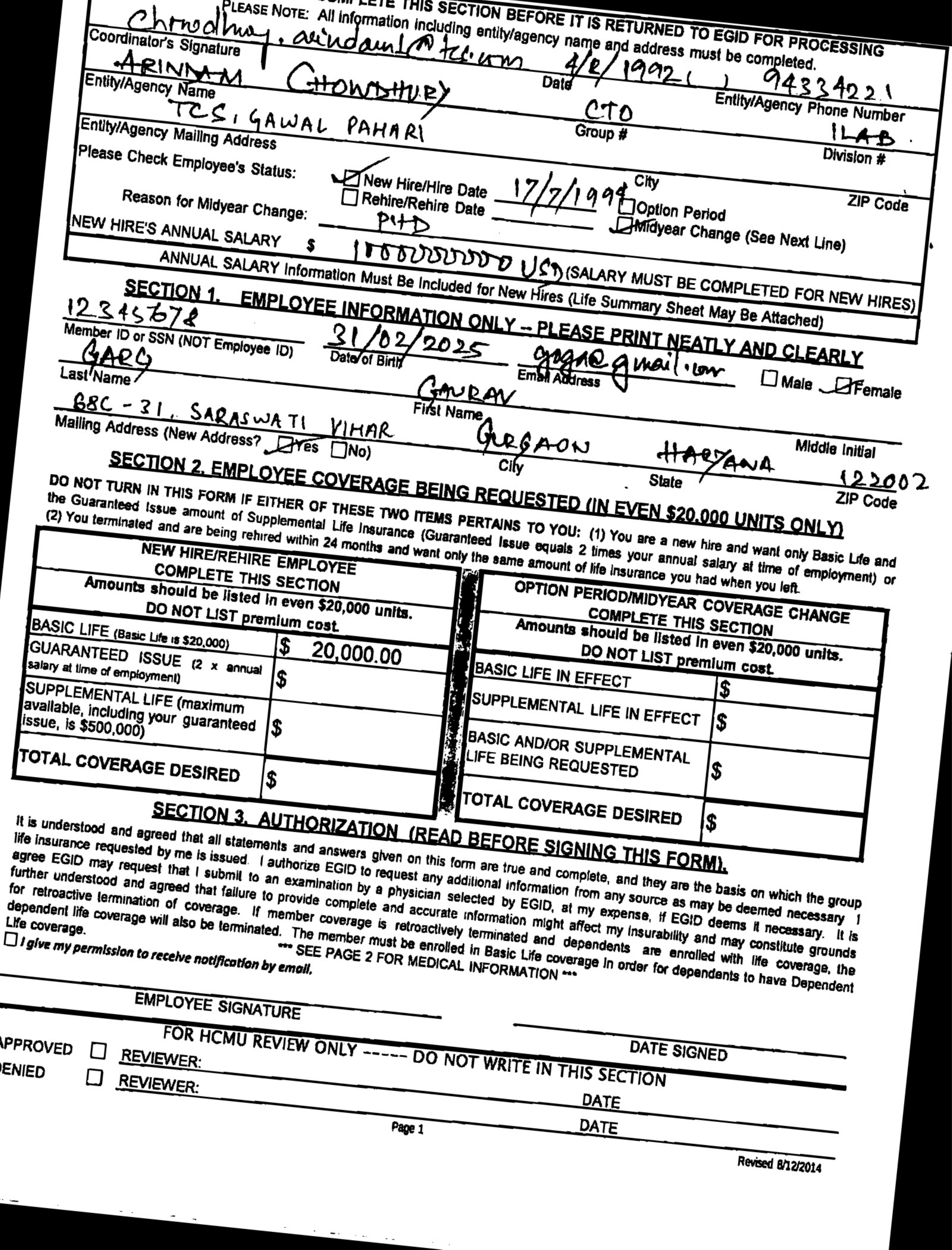}}
\subfigure[]{\includegraphics[width=0.23\linewidth]{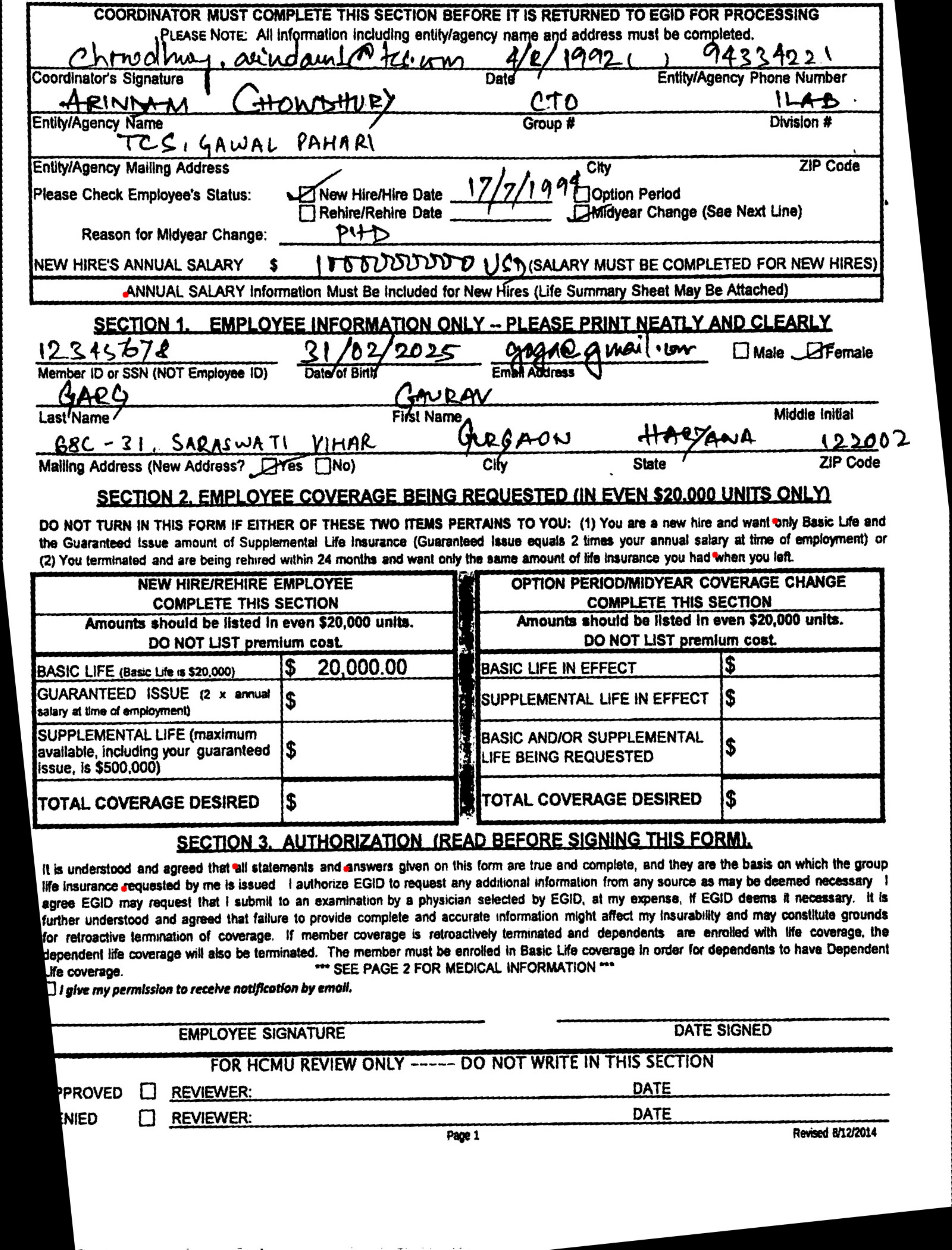}}
\subfigure[]{\includegraphics[width=0.23\linewidth]{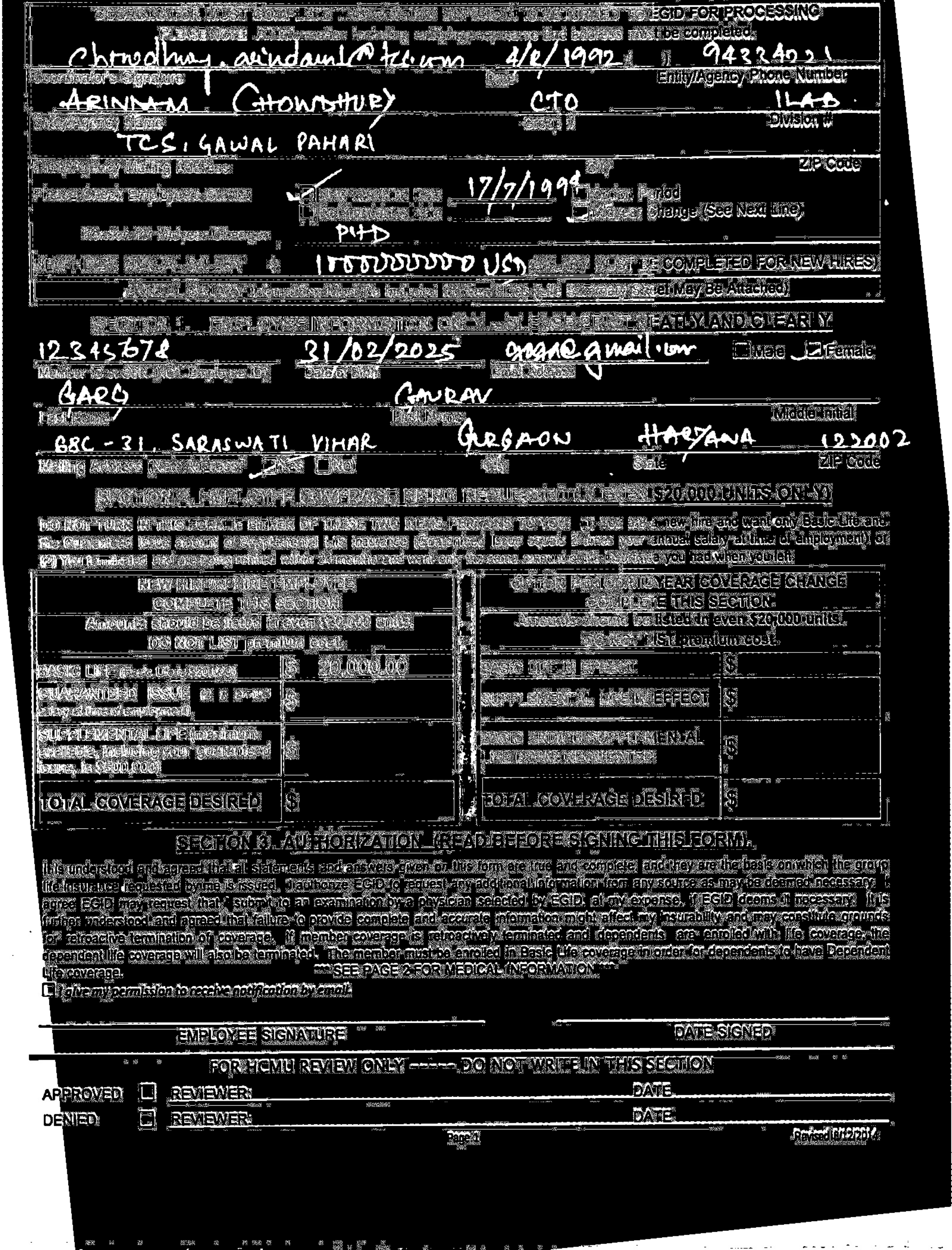}}
\end{center}
\vspace{-2mm}
\caption{(a), (b), (c) and (d) are the template, test image, aligned image, and the result of XOR operation between the template and aligned images for a sample document of the first dataset. (e), (f), (g) and (h) are the corresponding images for a sample document of the second dataset. The XOR operation allows us to visualize how perfectly the test document is aligned with the template, and the filled text in the test document stands out distinctly in bright white.}

\label{fig:qualitative_analysis}
\end{figure}

\renewcommand{\tabcolsep}{0.5mm}
\begin{table}[h]
\begin{center}
\vspace{-2mm}
\caption{Character recognition accuracy for fields in the second insurance dataset of application forms. Column (a) reports the accuracy for handwritten text tested on the HTR model given by Arindam et al., while column (b) gives the accuracy of handwritten text using the Google Vision API. This dataset does not contain added text in printed form.}
\vspace{1mm}
\begin{tabular}{|c|c|c|}
\hline
\textbf{Field} & {\textbf{HTR Model~\cite{chowdhury2018efficient} (A)}} & 
\textbf{Google Vision API (B)} \\
\hline
\textbf{Agency Name} & 78.7\% & 83.5\%\\
\textbf{Agency Address} & 78.3\% & 84.6\%\\
\textbf{First Name} & 80.1\% & 84.5\%\\
\textbf{Last Name} & 80.7\% & 86.7\%\\
\textbf{Applicant Address} & 78.4\% & 82.6\%\\
\textbf{City} & 81.9\% & 93.5\%\\
\textbf{State} & 83.2\% & 89.6\%\\
\hline
\end{tabular}
\end{center}
\label{tb:life_insurance_character}
\end{table}

\section{Experimental Results and Discussion}
\label{sec:experimental-results}
In this section, we present our experimental results on two document datasets of insurance claim forms. We used a threshold of 170, which was determined empirically, for binarization of documents.
Figure~\ref{fig:qualitative_analysis} shows some test documents and their corresponding documents aligned with the template. The fourth column in Figure~\ref{fig:qualitative_analysis} is obtained when we perform XOR operation between the aligned image and the template. It provides us greater visual understanding of how our system performs on the homography estimation and alignment task.

Alignment is followed by text field retrieval and classification of the text into printed or handwritten. We train a 5-layer CNN on patches of printed text cropped from text lines detected by CTPN~\cite{tian2016detecting}, and patches of handwritten text obtained from the IAM dataset~\cite{marti2002iam}. We obtain a test accuracy of 98.5\% when the model is tested on fields extracted from our documents. The quantitative measure of our information extraction pipeline is the character recognition accuracy of the retrieved text fields. Different models are employed for handwritten and printed text as specified in Section~\ref{sec:proposed-approach}. Table~\ref{tb:nationwide_character} reports the accuracies of some of the fields of interest in the first insurance dataset.

To get an estimate of the amount of perturbations that our system can handle, we make use of the second insurance dataset mentioned in Section~\ref{sec:dataset} and perform varying degrees of transformations like rotation, translation and scaling. We observe that our algorithm is able to handle translations and scaling of the test documents. For rotations, the system performance is unaffected for rotations upto $\pm7\degree$ in the x-y plane of the image. For rotations beyond this range, Tesseract output degrades significantly and thus, the image may not be aligned well. Horizontal and vertical translations range in between $\pm40\%$ of the document width and height respectively. Scaling factors largely depend on the font size on the document and the system performance is not impacted until the image gets pixelated. For our datasets, scaling works perfectly when the width and height are varied from 50\% to 200\% of their original values. The character recognition accuracies for the fields extracted during this stress test for the second insurance dataset are mentioned in Table~\ref{tb:life_insurance_character}. 

\section{Conclusion}
\label{sec:conclusion}
We proposed a character keypoint-based approach for homography estimation using textual information present in the document to address the problem of image alignment, specifically for scanned textual document images. Since such documents do not have smooth pixel intensity gradients for warp estimation, we cannot use the contemporary machine learning and deep learning algorithms relying on pixel intensity values for image alignment. To address these limitations, we create an automated system which takes an empty template document image and the corresponding filled test document, and aligns the test document with the template for extraction and analysis of textual fields. Experiments conducted on two real world datasets of insurance forms support the viability of our proposed approach.\vspace{1mm}


\bibliographystyle{IEEEtran}
\bibliography{mybib}

\end{document}